\documentclass[10pt,twocolumn,letterpaper]{article}

\usepackage{cvpr}              
\usepackage{multirow}
\usepackage{graphicx}
\usepackage{subfloat}

\usepackage[dvipsnames]{xcolor}

\usepackage[font=footnotesize,labelfont=bf]{caption}

\definecolor{cvprblue}{rgb}{0.21,0.49,0.74}
\usepackage[pagebackref,breaklinks,colorlinks,citecolor=cvprblue]{hyperref}
\usepackage[accsupp]{axessibility}
\usepackage{subcaption}
\usepackage{booktabs}

\def\paperID{8557} 
\def\confName{CVPR}
\def\confYear{2024}

\title{\emph{Light the Night}: A Multi-Condition Diffusion Framework for Unpaired Low-Light Enhancement in Autonomous Driving}

\author{Jinlong Li$^{1}$\thanks{Equal contribution. $^\dagger$Co-corresponding author, email address: xrxx3386@ucla.edu, h.yu19@csuohio.edu.
}
, Baolu Li$^{1*}$, Zhengzhong Tu$^{2}$, Xinyu Liu$^{1}$, Qing Guo$^{3}$,  Felix Juefei-Xu$^{4}$, \\ 
Runsheng Xu$^{5\dagger}$,  Hongkai Yu$^{1\dagger}$
\\ $^{1}$Cleveland State University  \hspace{2mm}   $^{2}$ University of Texas at Austin \\  $^{3}$Centre for Frontier AI Research (CFAR), A*STAR 
\hspace{2mm}  $^{4}$New York University 
\\  $^{5}$University of California, Los Angeles 
}

\begin{document}
\maketitle

\begin{abstract}
Vision-centric perception systems for autonomous driving have gained considerable attention recently due to their cost-effectiveness and scalability, especially compared to LiDAR-based systems. However, these systems often struggle in low-light conditions, potentially compromising their performance and safety. To address this, our paper introduces LightDiff, a domain-tailored framework designed to enhance the low-light image quality for autonomous driving applications. Specifically, we employ a multi-condition controlled diffusion model. LightDiff works without any human-collected paired data, leveraging a dynamic data degradation process instead. It incorporates a novel multi-condition adapter that adaptively controls the input weights from different modalities, including depth maps, RGB images, and text captions, to effectively illuminate dark scenes while maintaining context consistency. Furthermore, to align the enhanced images with the detection model's knowledge, LightDiff employs perception-specific scores as rewards to guide the diffusion training process through reinforcement learning. Extensive experiments on the nuScenes datasets demonstrate that LightDiff can significantly improve the performance of several state-of-the-art 3D detectors in night-time conditions while achieving high visual quality scores, highlighting its potential to safeguard autonomous driving.
\end{abstract}

\section{Introduction}

\begin{figure}[htb]
        \centering
	\begin{minipage}[b]{0.5\textwidth}
		\includegraphics[width=0.94\textwidth]{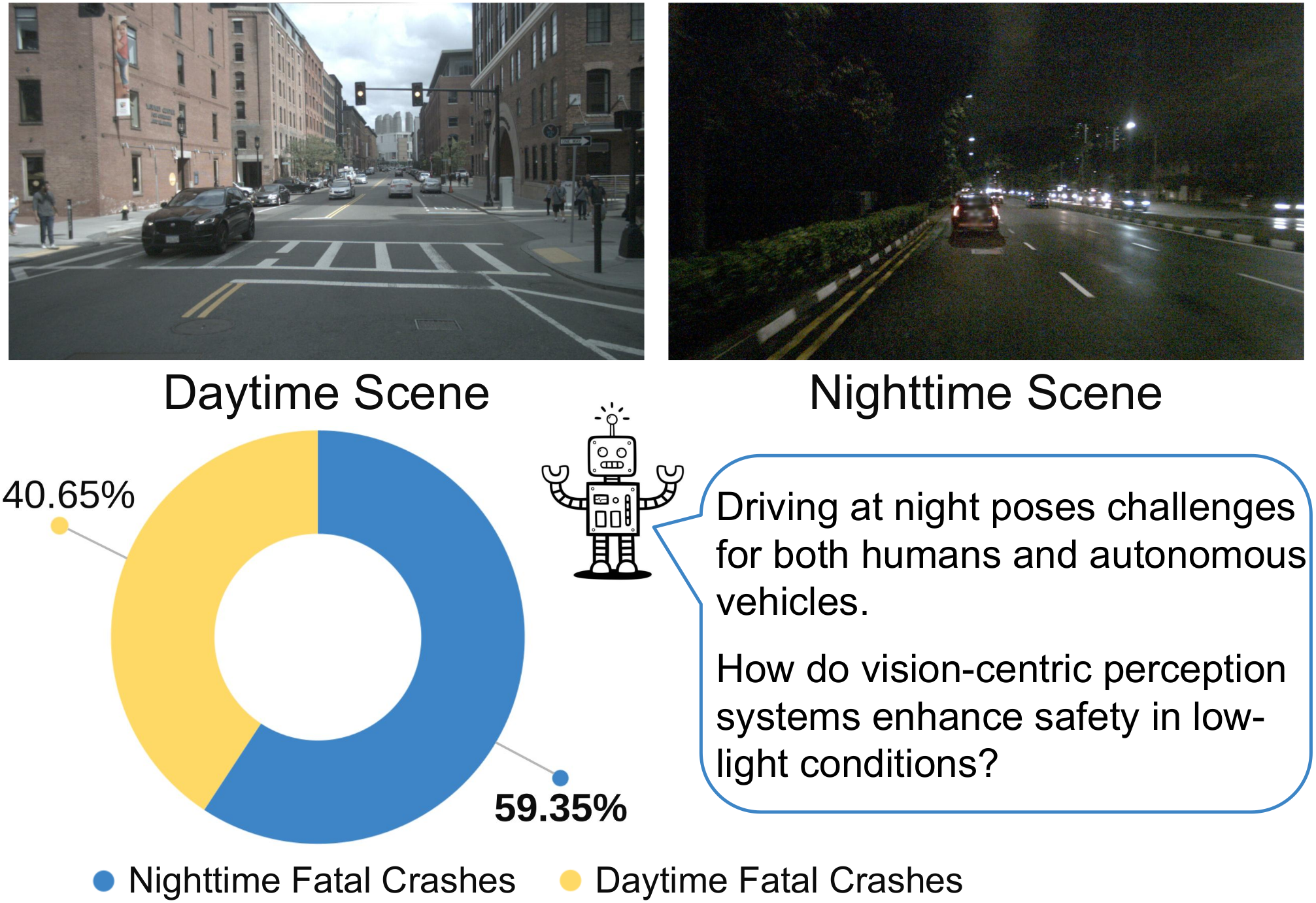}
	\end{minipage}
     \caption{\textbf{Nighttime driving scenarios pose a greater fatal threat than daytime.} The fatal rate at night is much  higher~\cite{ashraf2019catastrophic}. This paper aims to enhance nighttime images to improve the overall driving safety at night.}
	\label{fig:motivation} 
\end{figure}

Driving at night is challenging for humans, even more so for autonomous vehicles, as shown in Fig.~\ref{fig:motivation}. On March 18, 2018, a catastrophic incident highlighted this challenge when an Uber Advanced Technologies Group self-driving vehicle struck and killed a pedestrian in Arizona~\cite{ntsb_uber_2018}. This incident, resulting from the vehicle's failure to detect the pedestrian under low-light conditions accurately, has brought the safety concerns of autonomous vehicles to the forefront, especially in such demanding environments. As vision-centric autonomous driving systems predominantly relying on camera sensors become more prevalent, addressing the safety implications of low-light conditions has become increasingly critical to ensure the overall safety of these vehicles.

One intuitive solution is to collect extensive night-time driving data. However, this approach is not only labor-intensive and costly, but it also risks impairing daytime model performance due to the differing image distributions between night and day. To navigate these challenges, we propose a Lighting Diffusion (LightDiff) model, a novel method that eliminates the need for manual data collection and maintains model performance during the daytime.

LightDiff aims to enhance low-light camera images, improving perception model performance. Utilizing a dynamic low-light-degradation process, LightDiff generates synthetic day-night image pairs from existing daytime data to train. We then employ Stable Diffusion~\cite{rombach2022high} for its ability to produce high-quality visuals, effectively transforming night-time scenes into daytime equivalents. However, maintaining semantic consistency is critical in autonomous driving, a challenge with the original Stable Diffusion model. To overcome this, LightDiff incorporates multiple input modalities, such as estimated depth maps and camera image captions, coupled with a multi-condition adapter. This adapter intelligently determines the weighting of each input modality, ensuring the semantic integrity of the transformed images while keeping high visual quality. To guide the diffusion process not only to a direction that is visually brighter for humans but also for the perception model, we further finetune our LightDiff using reinforcement learning with perception-tailored domain knowledge in the loop. We conduct extensive experiments on the autonomous driving dataset nuScenes~\cite{caesar2020nuscenes} and demonstrate that our LightDiff can significantly improve 3D vehicle detection Average Precision~(AP) at nighttime by 4.2\% and 4.6\%, for two state-of-the-art models, BEVDepth~\cite{li2022bevdepth} and BEVStereo~\cite{li2022bevstereo}, respectively. Our contributions are summarized as follows:
\begin{itemize}
    \item We propose the Lighting Diffusion (LightDiff) model to enhance low-light camera images for autonomous driving, mitigating the need for extensive nighttime data collection and preserving daytime performance.
    
    \item  We integrate multiple input modalities including depth maps and image captions with a proposed multi-condition adapter to ensure semantic integrity in image transformation while maintaining high visual quality. We employ a practical process that generates day-night image pairs from daytime data for efficient model training. 
    
    \item We present a fine-tuning mechanism for LightDiff using reinforcement learning, incorporating perception-tailored domain knowledge (trustworthy LiDAR and statistical distribution consistency) to ensure that the diffusion process benefits both human visual perception and the perception model.

    \item Extensive experimentation with the nuScenes dataset demonstrates that LightDiff significantly improves 3D vehicle detection during the night and outperforms other generative models on multiple visual metrics.
\end{itemize}

\begin{figure*}[htb]
	\begin{minipage}[b]{1\textwidth}
		\centering
		\includegraphics[width=0.96\textwidth]{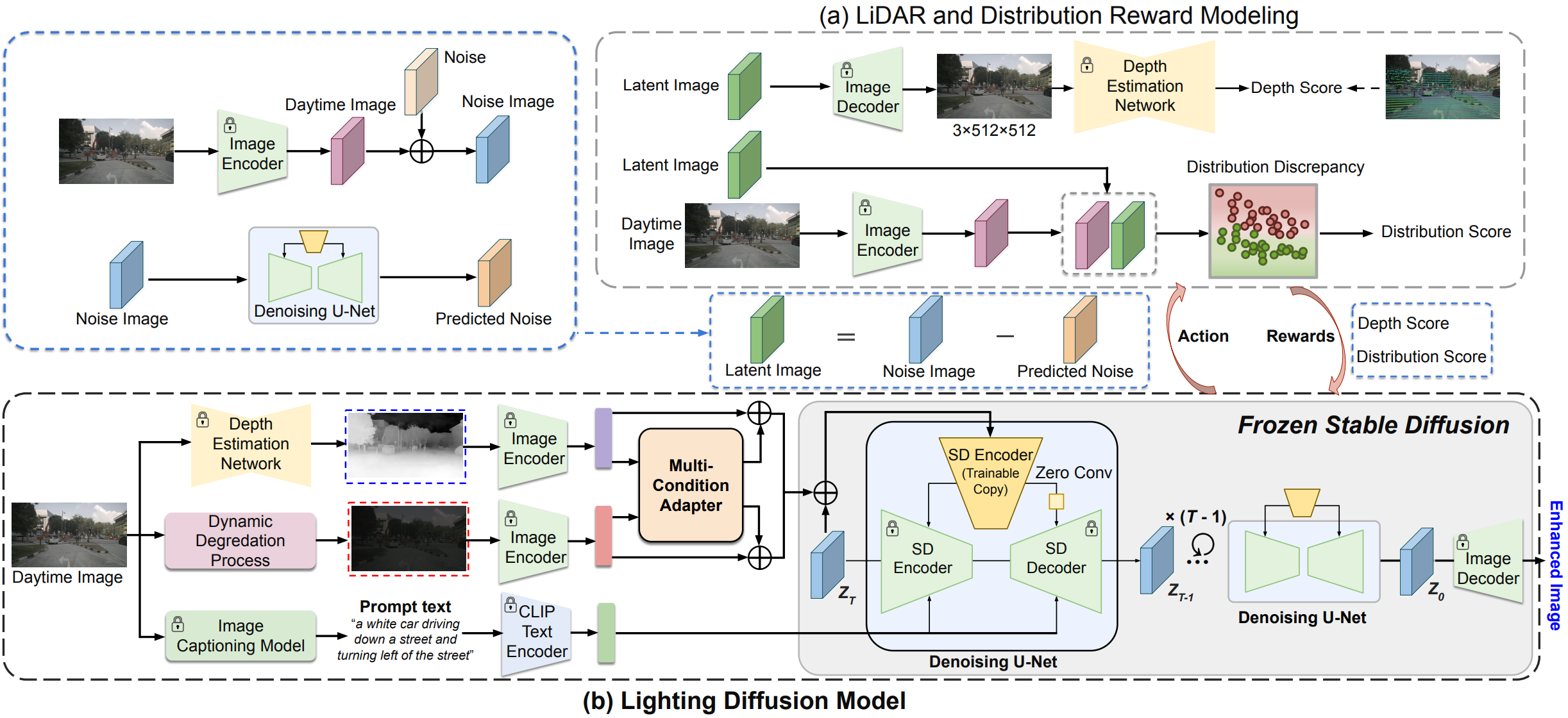}
	\end{minipage}
    \caption{The architecture of our Lighting Diffusion model (LightDiff). During the training stage, a Training Data Generation pipeline enables the acquisition of triple-modality data without any human-collected paired data. Our LightDiff employs a Multi-Condition Adapter to dynamically weight multiple conditions, coupled with LiDAR and Distribution Reward Modeling (LDRM), allowing for perception-oriented control.}
	\label{fig:proposed_training} 
\end{figure*}

\section{Related Work}~\label{Sec:Related Work}
\textbf{Dark Image Enhancement.} Dark image enhancement aims to  improve the visual quality and perceptibility of images suffering from dark conditions. It includes supervised methods~\cite{park2018distort, ren2019low}  that use paired datasets and unsupervised approaches~\cite{guo2020zero, li2021learning, liu2021retinex, ma2022toward} that enhance images without such paired data.  Some enhancement methods~\cite{huang2022deep, cai2018learning,jin2022unsupervised, jiang2021enlightengan,hu2021two} are developed to overcome the limitations in processing underexposed and/or overexposed regions in low-light conditions. There are some diffusion models for low-light image enhancement~\cite{yi2023diff,wang2023exposurediffusion,guo2023shadowdiffusion,fei2023generative}, which explicitly integrate degradation prior and diffusion generative capability, but they require paired data in training.

\noindent \textbf{Large Language Model in Vision.} Vision \& Language (VL) models~\cite{radford2021learning, goel2022cyclip, li2022blip, dou2022empirical, li2021align, alayrac2022flamingo,abdelfattah2023cdul} have shown obvious progress in computer vision.  CLIP~\cite{radford2021learning} acquires transferable visual concepts by natural language processing based supervision, learning knowledge from a large-scale dataset of 
image-caption pairs. Assisted by the language models, text/caption can be used to promote diverse computer vision tasks, such as CyCLIP~\cite{goel2022cyclip} and unCLIP~\cite{ramesh2022hierarchical}. Because the VL models contain substantial visual and language understanding, they can be utilized to evaluate image quality~\cite{zhang2023blind}. This insight inspires us to leverage VL model related techniques for enhancing low-light images~\cite{goel2022cyclip,kim2022diffusionclip,ramesh2022hierarchical}.

\noindent \textbf{Diffusion-based Generative Models.} The diffusion-based model~\cite{ho2020denoising} has achieved significant success in image synthesis through an iterative denoising process. Different  diffusion-based methods have been developed for the text-to-image generation task~\cite{gu2022vector,saharia2022photorealistic,ruiz2023dreambooth,zhang2023adding}, with outstanding performance in computer vision. Unlike some diffusion based methods relying on text prompts like Dreambooth~\cite{ruiz2023dreambooth}, the recent  ControlNet~\cite{zhang2023adding} incorporates the spatial condition based control signals into the pre-trained text-to-image diffusion models. Using the strong Stable Diffusion~\cite{rombach2022high} model as backbone which conducts the denoising process in the latent feature space, this paper makes efforts to enhance the dark visibility and address perception concerns to enhance the safety for driving at night.

\section{Methodology}~\label{Sec:Method}
We aim to propose a general framework for low-light image enhancement that can benefit the perception in autonomous driving.
To handle diverse driving-view scenarios, we exploit the strong generative prior imbued in the pre-trained Stable Diffusion model, which has been shown to deliver promising results for a variety of text-to-image and image-to-image tasks.
To train the model, we built a versatile nighttime image generation pipeline that can simulate realistic low-light images to produce training data pairs (as detailed in Sec.~\ref{sec:data_generation}).
Then, we introduce our proposed (\textbf{LightDiff}) model in Sec.~\ref{ssec:lightdiff}, a novel conditional generative model that can adaptively leverage various modalities of conditions---a low-light image, a depth map, and a text prompt---to predict the enhanced-light output.
Fig.~\ref{fig:proposed_training} depicts the entire pipeline of our proposed LightDiff architecture.
To improve our model's task awareness, we introduce a reward policy that considers guidance from trustworthy LiDAR and statistical distribution consistency, further described in Sec.~\ref{sssec:reward}.
Finally, we present a recurrent lighting  inference strategy to further boost the results of our model during test time, which is explained in Sec.~\ref{ssec:recurrent-infer}.

\subsection{Training Data Generation}~\label{sec:data_generation}
\begin{figure}[htb]
	\begin{minipage}[b]{0.5\textwidth}
		\includegraphics[width=0.94\textwidth]{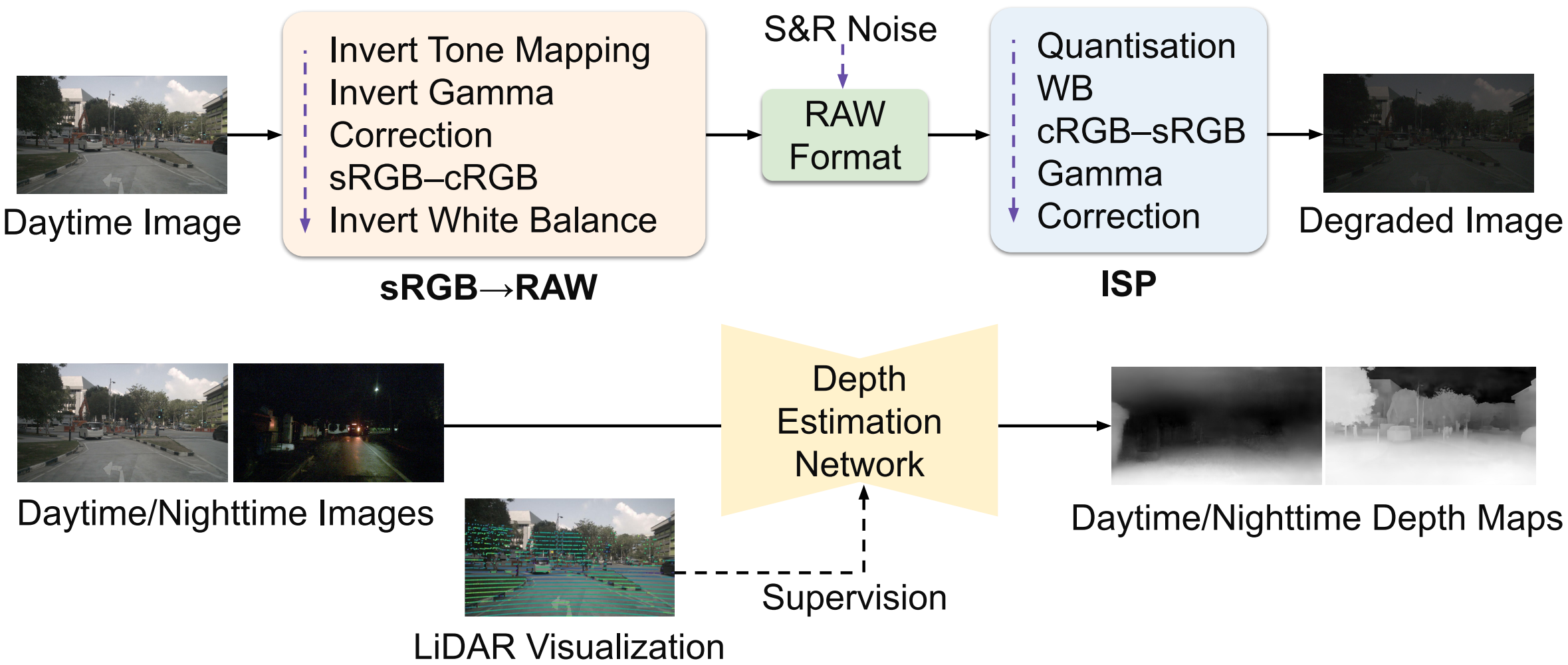}
	\end{minipage}
    \caption{The pipeline of our Training Data Generation. The low-light-degradation transform~\cite{cui2021multitask} is exclusively implemented during the training stage. The trained depth estimation network will be frozen to be used for the training and testing stages of our lighting diffusion model.}
	\label{fig:data_generation} 
\end{figure}

It is inherently challenging to collect nighttime-daytime paired images in dynamic driving scenarios.
In response to this challenge and to introduce more controlled conditions, we build a novel training data generation pipeline.
As illustrated in Fig.~\ref{fig:data_generation}, this pipeline generates multi-modality paired data, including \textit{1) instruction prompt}, \textit{2) trustworthy depth map generated by LiDAR}, and \textit{3) corresponding degraded dark light image}.
Starting with a daytime image $\text{I}_{day}$ as our target ground truth, we extract the text prompt by feeding it into a large image captioning model~\cite{berrios2023language}.
Meanwhile, we employ a pre-trained depth estimation network~\cite{ranftl2020towards} to obtain the corresponding depth map.
In common autonomous driving scenarios, where both LiDAR and camera sensors are provided, we project LiDAR point clouds onto the camera coordinate system as sparse points, which are then used as ground-truth supervision to train the depth estimation network. The pre-trained depth estimation network is frozen to be used for the training and testing of our lighting diffusion model. 
Unlike cameras, which are sensitive to illumination conditions, LiDAR maintains information consistency throughout both daytime and nighttime scenarios.
Drawing inspiration from~\cite{cui2021multitask}, we utilize a low-light-degradation transform $T_{deg}$ to synthesize vivid dark-light images $T_{deg}(\text{I}_{day})$, as depicted in Fig.~\ref{fig:data_generation}.
Specifically, we first transform the daytime image $\text{I}_{day}$ into RAW data using the sRGB $\to$ RAW process~\cite{brooks2019unprocessing}. Subsequently, we linearly attenuate the RAW image and introduce Shot and Read (S\&R) Noise, as commonly found in camera imaging systems~\cite{brooks2019unprocessing}. Finally, we apply the Image Signal Processing (ISP) pipeline to convert the low-light sensor measurement back to sRGB.
The overall low-light-degradation transform $T_{deg}$ can be simplified as:
\begin{align}
    T_{deg}(\text{I}_{day}) = T_{ISP}(T_{sRGB \to RAW}(\text{I}_{day}) + \text{I}_{noise}), 
    \label{eq:degrading}
\end{align}
which generates the degraded image  $\text{I}_{deg}$ similar to the dark nighttime image. We design a  Dynamic Degradation Process adopting  an online way  with randomized combinations of parameter ranges of Eq.~(\ref{eq:degrading}) to simulate the wider night driving scenes.

\subsection{Lighting Diffusion Model  (LightDiff)}
\label{ssec:lightdiff}
Our objective is to generate a pixel-level enhanced image which meticulously refines local textures and accurately reconstructs the global geometric outline of light details, conditioned on the triplet of multi-modal input data generated using our data pipeline (Sec.~\ref{sec:data_generation}).
Unlike previous conditional generative models~\cite{rombach2022high,zhang2023adding}, which only conditions on a \textit{single} modality, such as depth map, canny edge, \etc, our method recognizes and integrates the distinct contributions of each type of input modality towards the generation of the final output.
Processed by the Image Encoder, the latent features from the degraded image  $\text{I}_{deg}$ and the depth map $\text{I}_{dep}$, denoted as $\text{F}_{deg} \in \mathbb{R}^{H \times W \times C} $ and $\text{F}_{dep} \in \mathbb{R}^{H \times W \times C}$ respectively, are fed into the proposed Multi-Condition Adapter (Sec.~\ref{sssec:adaptor}), which adaptively fuses multiple conditions based on the global contribution of each input modality.
We adopt the ControlNet architecture~\cite{zhang2023adding} to learn the fused extra conditioning using a trainable copy of the UNet encoder, while leaving the backbone diffusion model frozen.

\subsubsection{Preliminary: Stable Diffusion}
We employ Stable Diffusion (SD), a large-scale text-to-image pre-trained latent diffusion model to achieve dark enhancement in dynamic driving scenarios.
By definition, diffusion models generate data samples through a sequence of denoising steps that estimate the score of the data distribution.
For improved efficiency and stabilized training, SD pretrains a variational autoencoder (VAE)~\cite{kingma2013auto} that compresses an image \(I\) into a latent \(z\) with encoder \(E\) and reconstructs it with decoder \(\mathcal{D}\).
Both the diffusion and denoising processes happen in the latent space.
In the diffusion process, Gaussian noise with variance \(\beta_t \in (0, 1)\) at time \(t\) is added to the encoded latent \(z = E(I)\) to produce the noisy latent:
\begin{equation}
z_t = \sqrt{\bar{\alpha}_t}z + \sqrt{1-\bar{\alpha}_t}\chi,
\end{equation}
where \(\chi \sim \mathcal{N}(0, \textbf{I})\), \(\alpha_t = 1 - \beta_t\), and \(\bar{\alpha}_t = \prod^{t}_{s=1}\alpha_s\). When \(t\) is sufficiently large, the latent \(z_t\) approximates a standard Gaussian distribution.
A network \(\epsilon_{\theta}\) is learned by predicting the noise \(\epsilon\) conditioned on \(c_t\) (text prompts) at a randomly chosen time-step \(t\). The optimization objective of the latent diffusion model is defined as:
\begin{equation}
\mathcal{L}_{LDM} = \mathbb{E}_{z, c_t, t, \epsilon}[||\epsilon-\epsilon_{\theta}(z_t, c_t, t)||^2_2],
\end{equation}
where \(t\) is uniformly sampled and \(\epsilon\) is sampled from the Gaussian distribution.

\subsubsection{Multi-Condition Adapter}
\label{sssec:adaptor}

To discern the significance of different visual conditions, we introduce a novel multi-condition adapter, that is designed to dynamically weighs the conditions based on input data.
Particularly, the latent features of the dark-light input $\text{F}_{deg} \in \mathbb{R}^{H \times W \times C}$ with the paired depth map $\text{F}_{dep} \in \mathbb{R}^{H \times W \times C}$ are concatenated as $\mathbf{F}_{(dep, deg)}$ and fed into a convolution layer.
It is then reshaped to $\mathbb{R}^{2C \times (H \times W)}$ denoted as $\mathbf{F}^{c}_{(dep, deg)}$.
A softmax layer is applied to the matrix multiplication of $\mathbf{F}^{c}_{(dep, deg)}$ and its transpose, obtaining the multi-condition weights $\mathbf{W} \in \mathbb{R}^{2C \times 2C}$:
\begin{equation}
w_{(dep, deg)} = \frac{\exp(\mathbf{F}^{c}_{dep} \cdot \mathbf{F}^{c}_{deg})}{\sum_{c}   \exp(\mathbf{F}^{c}_{dep} \cdot \mathbf{F}^{c}_{deg})},
\end{equation}
where $w_{(dep, deg)}$ measures the impact of $\text{F}_{deg}$ on $\text{F}_{dep}$.
The transposed $\mathbf{W}$ is multiplied with $\mathbf{F}^{c}_{(dep, deg)}$, then reshaped to $\mathbb{R}^{2C \times H \times W}$.
An element-wise sum operation with $\mathbf{F}^{c}_{(dep, deg)}$ yields the   output $ \text{F}^{'}_{deg} \in \mathbb{R}^{2C \times H \times W}$:
\begin{equation}
\text{F}^{'}_{deg} = \sum_{c} (w_{(dep, deg)} \mathbf{F}^{c}_{dep}) + \mathbf{F}^{c}_{deg}.
\end{equation}
In the same way, we could obtain $\text{F}^{'}_{dep} \in \mathbb{R}^{2C \times H \times W}$. The final output  represents a weighted combination of all the conditions, capturing semantic dependencies between the multiple  modalities. The multi-condition adapter is succinctly represented as:
\begin{equation}
\text{F}^{'}_{deg}, \text{F}^{'}_{dep} = \text{MC-Adaptor}(\text{F}_{deg}, \text{F}_{dep}).
\end{equation}

\subsubsection{Controlling the Stable Diffusion Model} 

Inspired by~\cite{zhang2023adding}, we employ an additional conditioning network trained from scratch to encode additional condition information.
We first use the encoder of Stable Diffusion's pre-trained VAE to map \(I_{deg}\) and \(I_{dep}\) into the latent space, obtaining the conditional latents \(\text{F}_{deg}\) and \(\text{F}_{dep}\).
The UNet denoiser in SD performs latent diffusion, which includes an encoder, a middle block, and a decoder.
We create an additional copy of the UNet encoder (denoted in orange in Fig.~\ref{fig:proposed_training}(b)) to inject additional visual conditions.
After being processed by the multi-condition adapter, the conditional latents \(\text{F}^{'}_{deg}\) and \(\text{F}^{'}_{dep}\) are concatenated with the randomly sampled noise \(z_t\) as inputs to the trainable copy of encoder.
Their outputs are added back to the original UNet decoder, with a \(1 \times 1\) convolutional layer (denoted as an orange rectangle in Fig.~\ref{fig:proposed_training}(b)) applied before the residual addition operation for each scale.
During finetuning, the additional module and these \(1 \times 1\) convolutional layers are optimized simultaneously.
The entire network \(\epsilon_{\theta}\) learns to predict the noise \(z_n\) added to the noisy image \(z_t\) by minimizing the following latent diffusion objective:
\begin{equation}
\mathcal{L}_\text{Lighting} = \mathbb{E}_{z_t, c_t, c_d, t, \epsilon}[||\epsilon-\epsilon_{\theta}(z_t, c_t, c_d, t)||^2_2],
\end{equation}
where \(c_d\) represents the condition combining the dark-light image and the depth map.

\subsection{LiDAR and Distribution Reward Modeling}
\label{sssec:reward}
To achieve fine-grained task-oriented control, we introduce a reward policy that considers guidance from trustworthy LiDAR and statistical distribution consistency during training our lighting diffusion model. 
We create a training schedule where the reward is applied only on predicted clean latent images $z_I$ when the sampled time step $t$ is less than a threshold $\tau$.
We leverage a frozen depth estimation network and apply a distribution-aware statistical consistency module to enforce distribution alignment.
As shown in Fig.~\ref{fig:proposed_training}, the $z_I$ is fed into the image decoder to generate the pixel-level image feature $I_{pred}$ with the same shape as the original real daytime image.
The depth estimation network predicts the depth map, whose misalignment metric ($L_{\text{Depth}}$) is the mean square error with the ground truth depth map by the trustworthy LiDAR point clouds.

To address the distribution gap between the enhanced lighting image and the real daytime image, we examine the relationship between statistical differences and feature distributions.
Previous studies~\cite{hou2023evading} have established a positive correlation between statistical differences and distribution disparities.
As such, to minimize the discrepancy of feature distributions between $z_I$ and $z_{gt}$, we introduce the distribution-aware statistical consistency module, utilizing the Maximum Mean Discrepancy (MMD)~\cite{JMLR:v13:gretton12a} distance ($L_{\text{MMD}}$) as a metric.
Specifically, let $\mathcal{Z}_{I} = \{z_{I}^{i}\}$ and $\mathcal{Z}_{gt} = \{z_{gt}^{i}\}$ represent a set of enhanced lighting and real daytime features, respectively.
The reward model takes the predicted clean latent image $z_I$ as input and outputs two scalar rewards, namely depth and distribution scores.
Following the Reinforcement Learning (RL) training strategy~\cite{ouyang2022training,le2022deep}, the agent, represented by the UNet denoiser $\epsilon_{\theta}$, presents a predicted clean latent image $z_I$ and expects a response based on $z_I$.
It takes the $z_I$ and produces a reward determined by the reward model, thus concluding the episode.
We minimize the following combined objective function in the RL training:
\begin{equation}
    \begin{aligned}
    \mathcal{L}_{\text{obj}} &= \mathbb{E}_{z_t, c_t, c_d, t, \epsilon}\left[||\epsilon-\epsilon_{\theta}(z_t, c_t, c_d, t)||^2_2\right] \\
    &+ \Phi^{RL}_{\mathbb{E}_{z_t, c_t, c_d, t, \epsilon}}\left(L_{\text{MMD}}(\mathcal{Z}_{I}, \mathcal{Z}_{gt}), L_{\text{Depth}}(z_I)\right),
    \end{aligned}
\end{equation}
where $\Phi^{RL}_{\mathbb{E}_{z_t, c_t, c_d, t, \epsilon}}$ is the learned policy. This designed reward modeling will guide the training of our lighting diffusion model by leveraging the trustworthy LiDAR and statistical distribution consistency.

\begin{figure}[htb]
        \includegraphics[width=0.95\columnwidth]{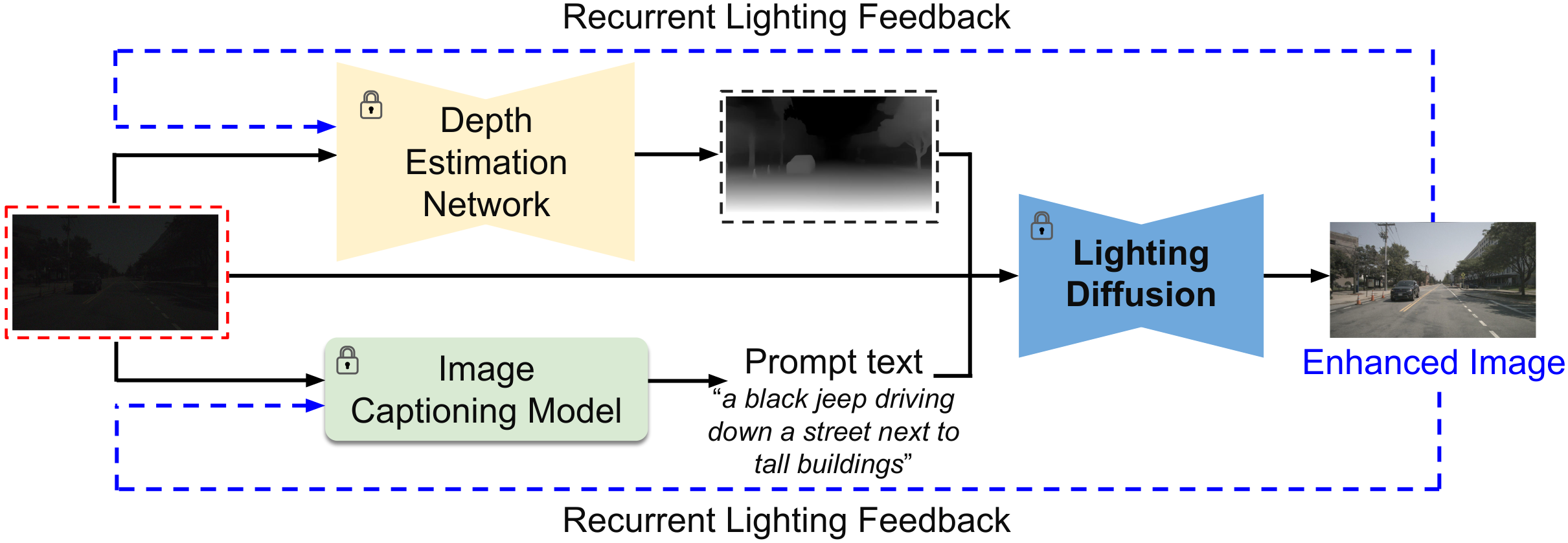}
    \caption{Illustration of the Recurrent Lighting Inference. It is designed to enhance the precision of generating text prompts and depth maps, thereby mitigating adverse effects on dark images.}
	\label{fig:inference} 
\end{figure}

\subsection{Recurrent Lighting Inference}
\label{ssec:recurrent-infer}
Real-night images, in contrast to clear daytime images, often suffer from low visibility and uneven light distribution.
These conditions pose significant challenges for depth generation of the pre-trained depth estimation network, as well as the image captioning model.
To address these issues, we implement an iterative feedback process that includes refining text prompts and tuning generated depth maps, as illustrated in Fig.~\ref{fig:enhanced modality}. 
This process, executed in a loop with the depth estimation network, image captioning model, and lighting diffusion model remaining constant, aims to improve the accuracy of text prompts and refine depth maps for initial dark images, thereby enhancing the overall lighting results.
Particularly, the procedure starts with feeding a real nighttime image into the depth estimation network and image captioning model to acquire an initial estimate of the corresponding text prompt and depth map.
These inputs are then employed by the lighting diffusion model to produce an enhanced lighting image.
Subsequently, we feed this initial enhanced image to replace the original nighttime image to further generate a refined text prompt and depth map, which are utilized as inputs for the next iteration.
The cycle repeats until the similarity of the final generated images stabilizes, but in practice we found only two iterations are sufficient to generate a high-quality enhanced image.

\begin{figure*}[htb]
        \centering
        \subfloat[Night Images]{%
        \includegraphics[width=0.35\columnwidth]{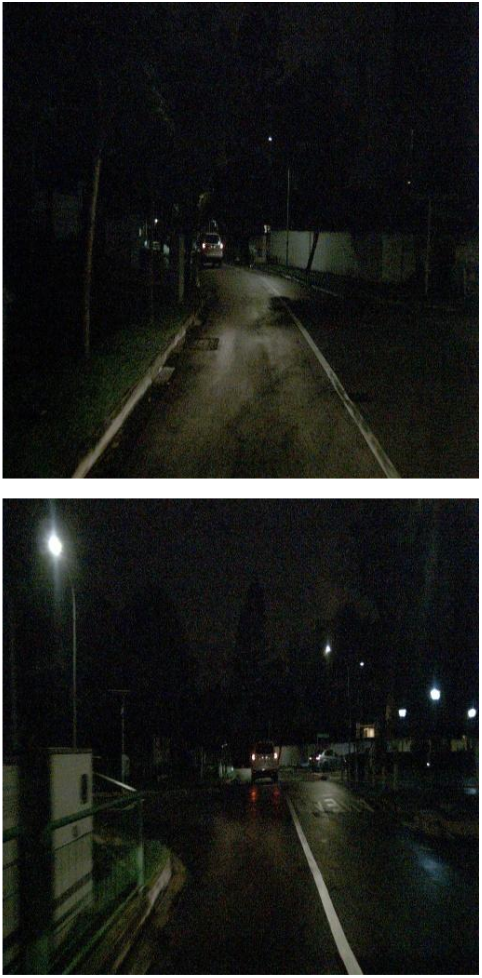}%
        }
        \subfloat[RUAS-LOL~\cite{liu2021retinex}]{%
        \includegraphics[width=0.35\columnwidth]{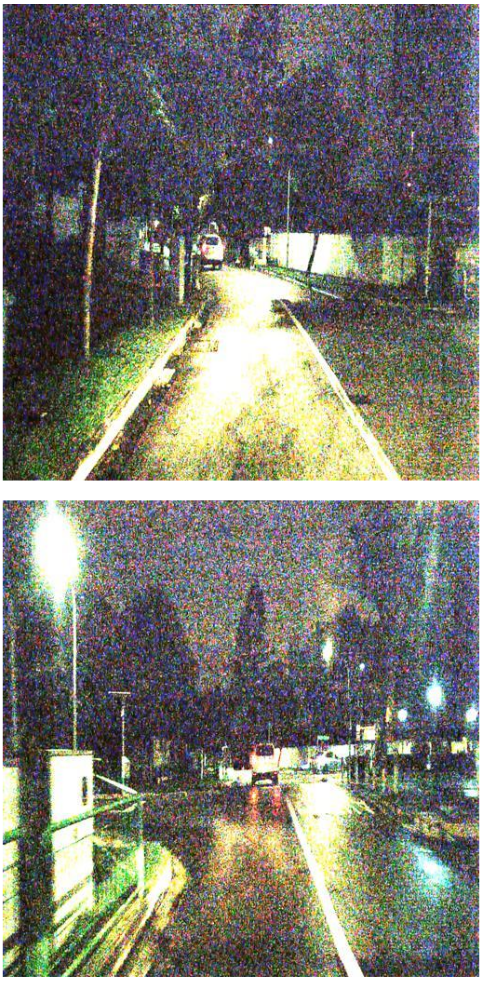}%
        }
        \subfloat[SCI-difficult~\cite{ma2022toward}]{%
        \includegraphics[width=0.35\columnwidth]{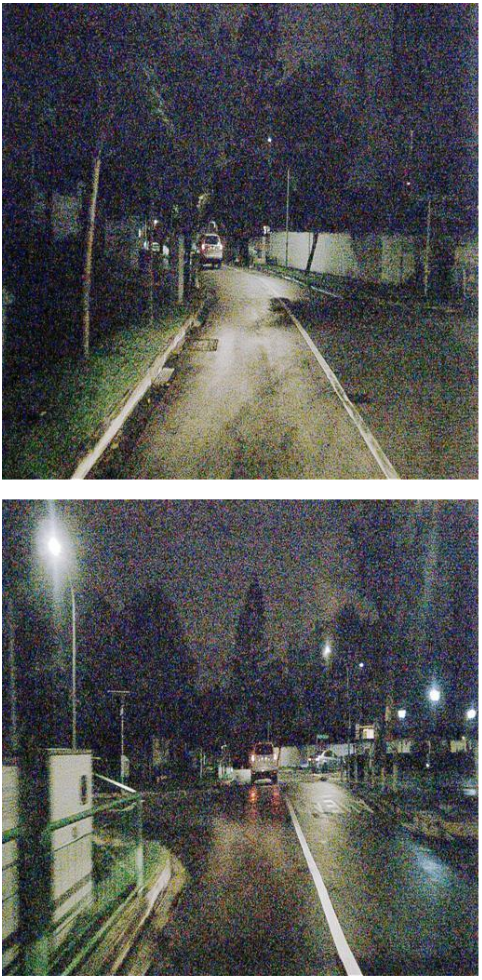}%
        }
        \subfloat[Zero-DCE++~\cite{li2021learning}]{%
        \includegraphics[width=0.35\columnwidth]{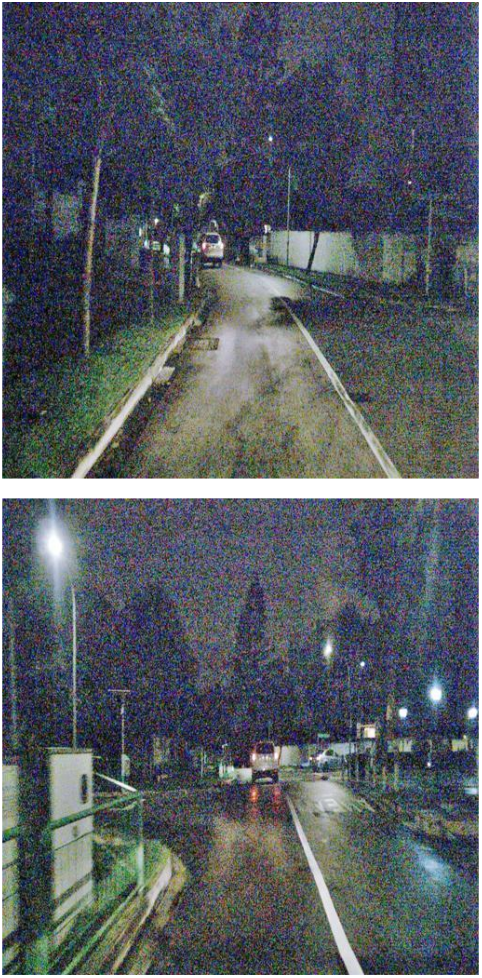}%
        }
        \subfloat[URetinex-Net~\cite{wu2022uretinex}]{%
        \includegraphics[width=0.35\columnwidth]{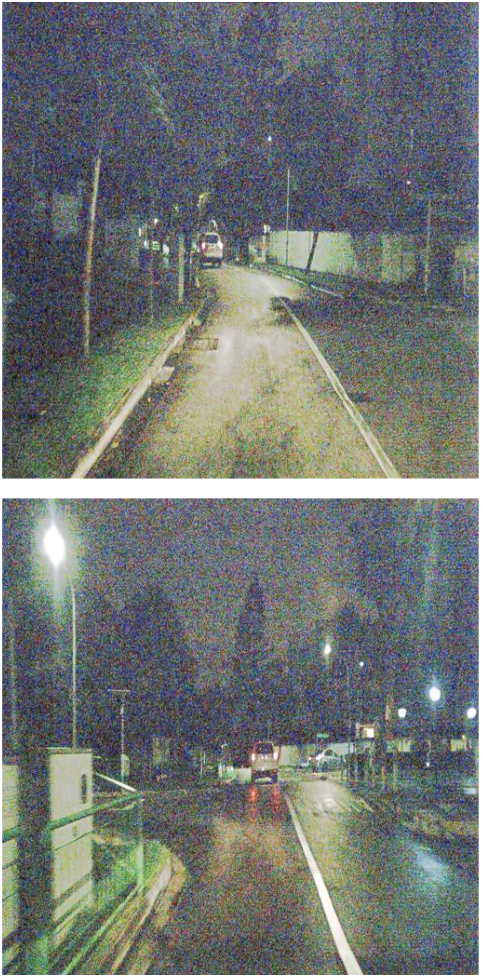}%
        }
        \subfloat[Ours]{%
        \includegraphics[width=0.35\columnwidth]{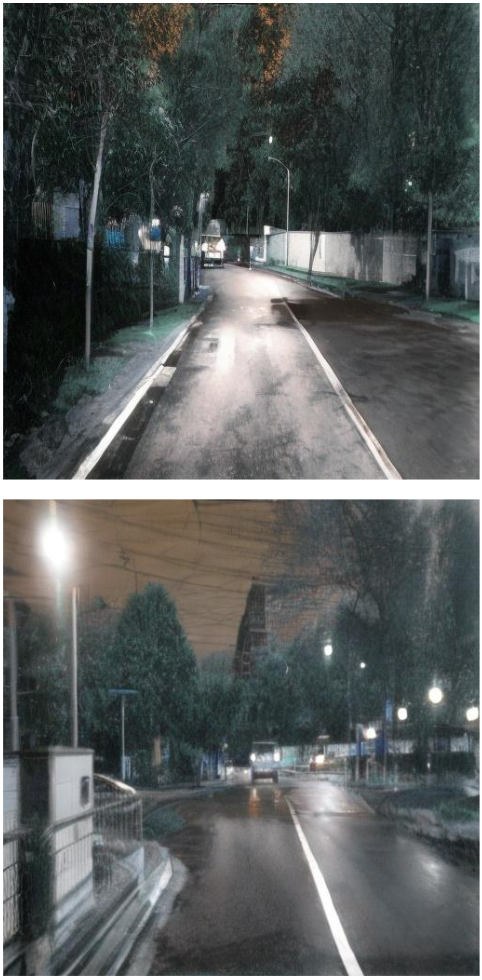}%
        }
    \caption{Visual comparison on the example nighttime images in the nuScenes validation set.}
	\label{fig:comparison_generating} 
\end{figure*}

\begin{figure*}[htb]
        \centering
        \subfloat[Night Images]{%
        \includegraphics[width=0.7\columnwidth]{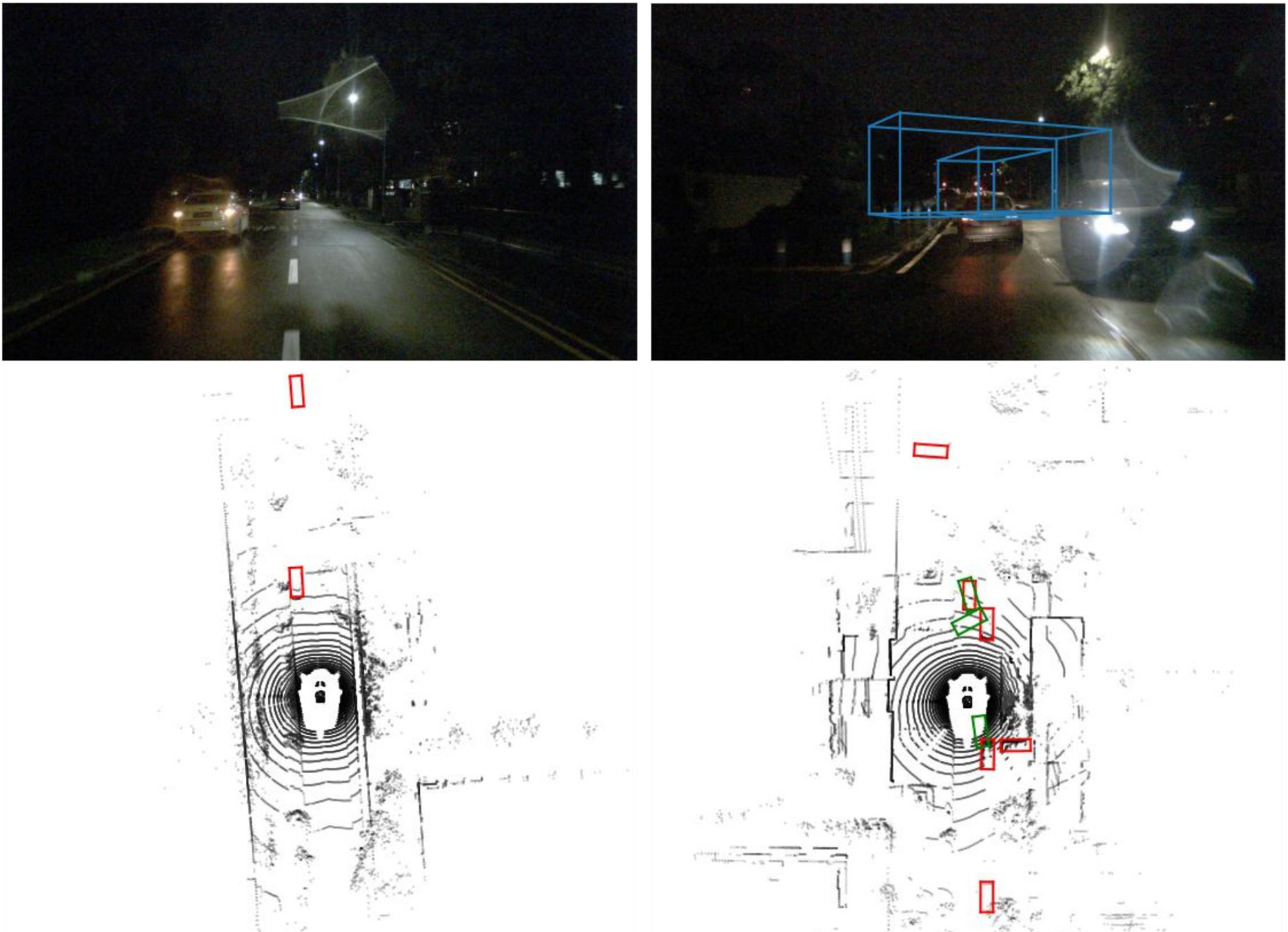}%
        }
        \subfloat[SCI-difficult~\cite{ma2022toward}]{%
        \includegraphics[width=0.7\columnwidth]{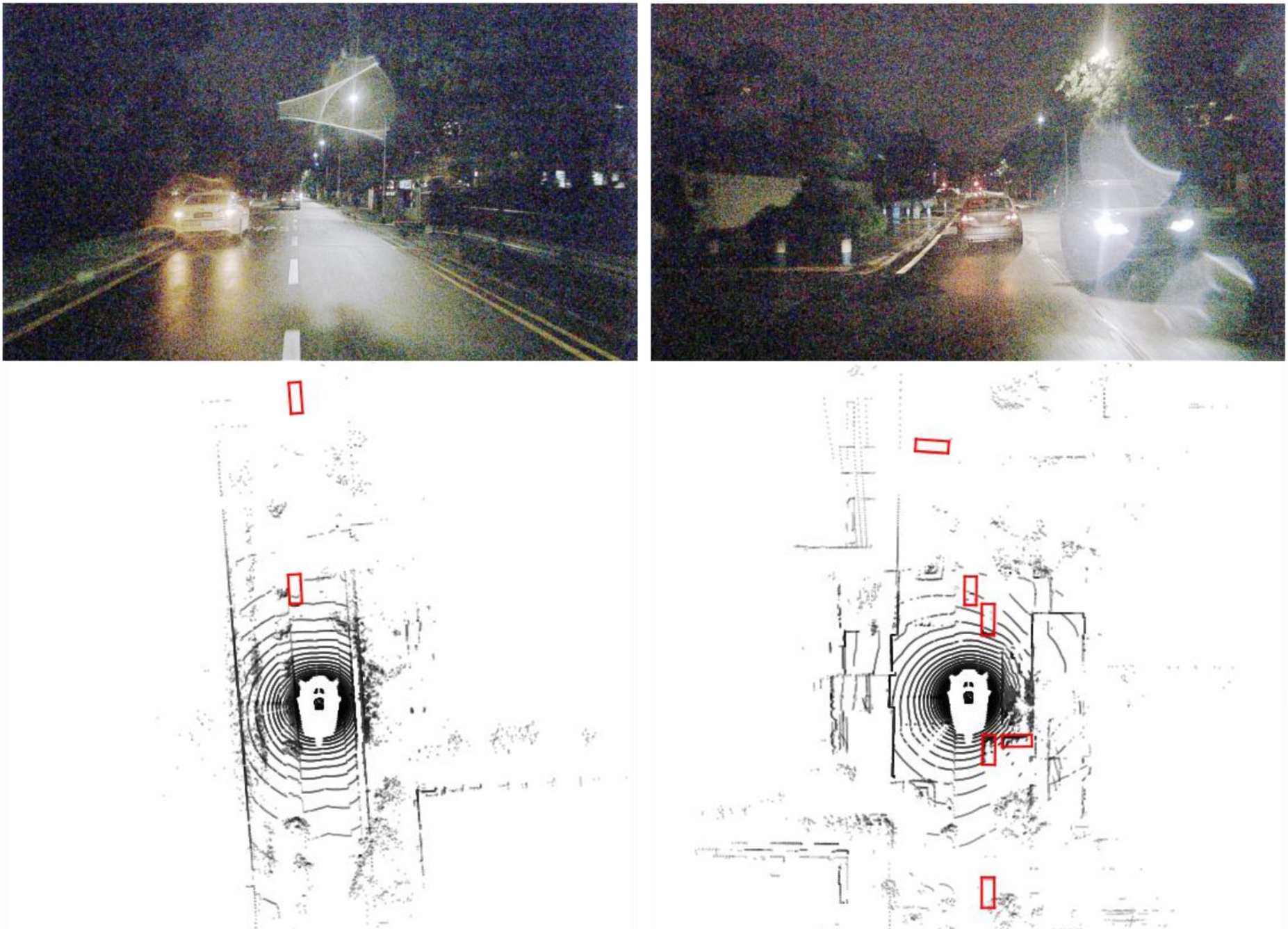}%
        }
        \subfloat[Ours]{%
        \includegraphics[width=0.7\columnwidth]{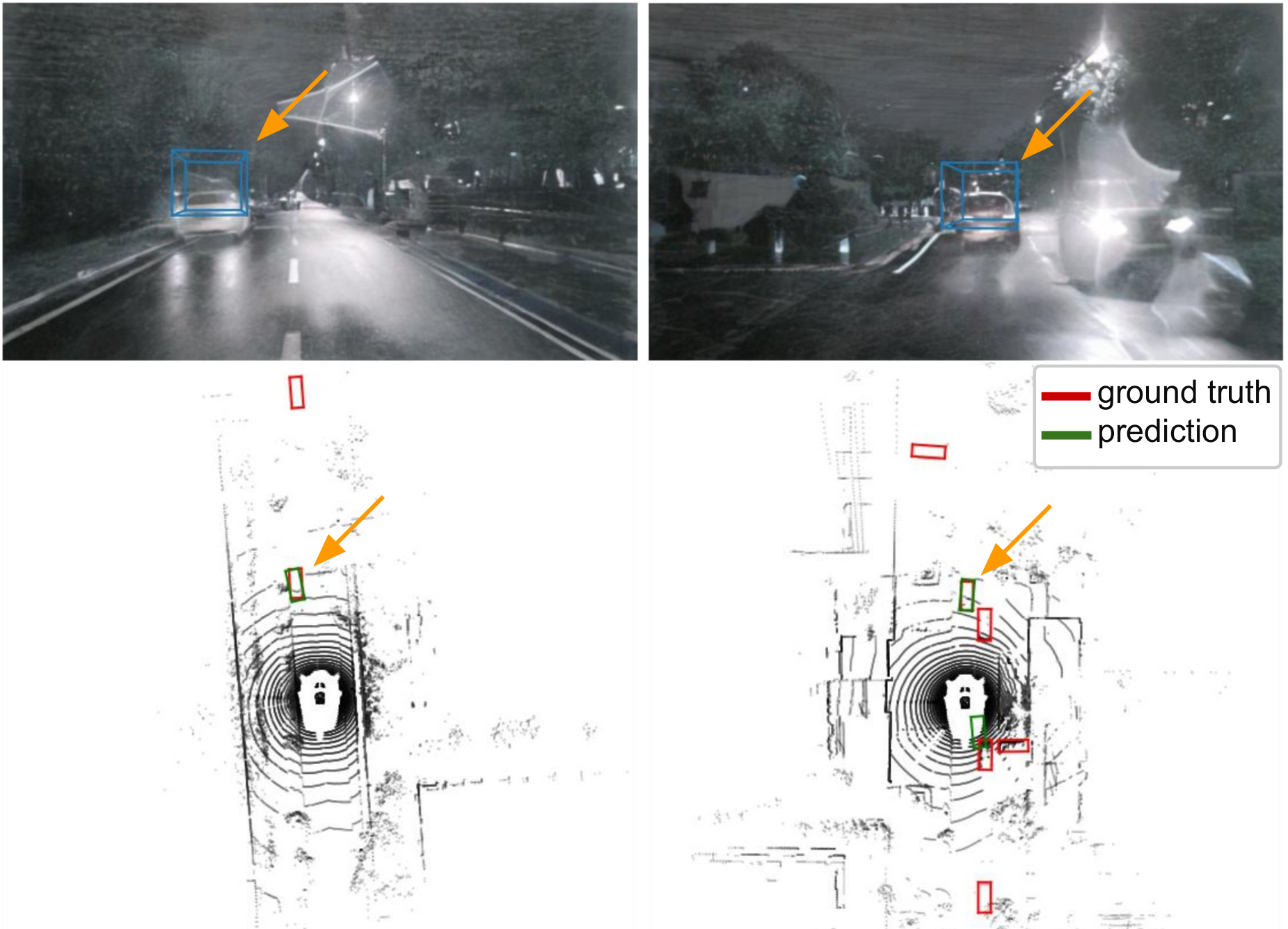}%
        }
    \caption{Visualization of 3D detection results on the example nighttime images in the nuScenes validation set. We employ BEVDepth~\cite{li2022bevdepth} as the 3D detector and visualize both the front view of camera and the Bird’s-Eye-View.}
	\label{fig:comparison_detection} 
\end{figure*}

\begin{table*}[htb]
\caption{Quantitative comparison of image quality on the nuScenes nighttime validation set. The best and second performance are marked in {\color[HTML]{FE0000} \textbf{red}} and {\color[HTML]{3531FF} \textbf{blue}}.}
\label{tab:vis_result}
\centering
\resizebox{0.86\textwidth}{!}{
\begin{tabular}{@{}c|c|c|c|c|c|c|c|c@{}}
\toprule
 &
  Methods &
  MUSIQ$\uparrow$ &
  HyperIQA$\uparrow$ &
  MANIQA$\uparrow$ &
  NIMA$\uparrow$ &
  TReS$\uparrow$ &
  ILNIQE$\downarrow$ &
  NIQE$\downarrow$ \\ \cmidrule(l){2-9} 
\multirow{-2}{*}{Type} &
  Night Image &
  16.295 &
  0.229 &
  0.034 &
  3.197 &
  20.753 &
  35.627 &
  3.700 \\ \midrule
 &
  Afifi et al.~\cite{afifi2021learning} &
  13.562 &
  0.207 &
  0.045 &
  3.715 &
  19.731 &
  32.207 &
  8.388 \\
 &
  URetinex-Net~\cite{wu2022uretinex} &
  24.086 &
  0.244 &
  0.065 &
  {\color[HTML]{3531FF} \textbf{4.503}} &
  53.571 &
  38.972 &
  4.460 \\
 &
  SNR-Aware-LOLv1~\cite{xu2022snr} &
  13.591 &
  0.229 &
  0.032 &
  3.947 &
  22.893 &
  {\color[HTML]{3531FF} \textbf{30.119}} &
  8.898 \\
 &
  SNR-Aware-LOLv2real~\cite{xu2022snr} &
  13.591 &
  0.230 &
  0.032 &
  3.947 &
  23.025 &
  {\color[HTML]{3531FF} \textbf{30.119}} &
  8.898 \\
 &
  SNR-Aware-LOLv2synthetic~\cite{xu2022snr} &
  14.528 &
  0.219 &
  0.026 &
  3.414 &
  12.571 &
  38.652 &
  8.967 \\ 
   &
  SNR-Aware-nuScene~\cite{xu2022snr} &
  16.928 &
  0.241 &
  0.056 &
  3.767 &
  24.013 &
  35.087 &
  7.664 \\ 
\multirow{-5}{*}{Supervised}  &
  ExposureDiffusion~\cite{wang2023exposurediffusion} &
  20.365 &
  0.204 &
  0.026 &
  3.977 &
  10.795 &
  50.562 &
  5.020 \\ 
  &
  ShadowDiffusion~\cite{guo2023shadowdiffusion} &
  {\color[HTML]{3531FF} \textbf{40.695}} &
  {\color[HTML]{FF0000} \textbf{0.446}} &
  0.078 &
  3.808 &
  {\color[HTML]{FF0000} \textbf{67.086}} &
  57.984 &
  5.607 \\ 
  
  \midrule
 &
  Zero-DCE~\cite{guo2020zero} &
  24.562 &
  0.250 &
  0.064 &
  4.360 &
  52.988 &
  35.804 &
  4.316 \\
 &
  Zero-DCE++~\cite{li2021learning} &
  20.276 &
  0.242 &
  0.059 &
  4.375 &
  49.553 &
  36.025 &
  4.232 \\
 &
  RUAS-LOL~\cite{liu2021retinex} &
  24.587 &
  0.257 &
  0.056 &
  4.311 &
  43.836 &
  52.761 &
  6.031 \\
 &
  RUAS-MIT5K~\cite{liu2021retinex} &
  14.965 &
  0.234 &
  0.041 &
  4.046 &
  34.945 &
  46.425 &
  4.506 \\
 &
  RUAS-DarkFace~\cite{liu2021retinex} &
  20.277 &
  0.248 &
  0.067 &
  4.382 &
  45.488 &
  55.351 &
  5.595 \\
 &
  SCI-easy~\cite{ma2022toward} &
  15.240 &
  0.250 &
  0.044 &
  4.004 &
  33.928 &
  31.679 &
  3.975 \\
 &
  SCI-medium~\cite{ma2022toward} &
  15.513 &
  0.241 &
  0.056 &
  4.374 &
  50.316 &
  36.317 &
  4.423 \\
 &
  SCI-difficult~\cite{ma2022toward} &
  34.718 &
  0.260 &
  {\color[HTML]{3531FF} \textbf{0.081}} &
  4.370 &
  58.037 &
  34.583 &
  5.050 \\
 &
  EnlightenGAN~\cite{jiang2021enlightengan} &
  20.686 &
  0.242 &
  0.070 &
  4.383 &
  42.829 &
  40.080 &
  4.307 \\
 &
  LESNet~\cite{jin2022unsupervised} &
  19.410 &
  0.205 &
  0.032 &
  3.477 &
  14.453 &
  32.784 &
  7.905 \\
\multirow{-11}{*}{Unsupervised} &
  CLIP-LIT~\cite{liang2023iterative} &
  23.805 &
  0.229 &
  0.064 &
  4.402 &
  49.557 &
  42.560 &
  4.701 \\ \midrule
 &
  SCI~\cite{ma2022toward} &
  14.781 &
  0.238 &
  0.044 &
  3.909 &
  34.819 &
  34.220 &
  4.118 \\
 &
  EnlightenGAN~\cite{jiang2021enlightengan} &
  16.334 &
  0.239 &
  0.035 &
  3.309 &
  24.654 &
  33.294 &
  {\color[HTML]{FF0000} \textbf{3.397}} \\
 &
  CLIP-LIT~\cite{liang2023iterative} &
  16.288 &
  0.229 &
  0.033 &
  3.206 &
  20.766 &
  35.681 &
  3.703 \\
\multirow{-4}{*}{\begin{tabular}[c]{@{}c@{}}Unsupervised\\ (retrained)\end{tabular}} &
  Ours &
  {\color[HTML]{FE0000} \textbf{51.674}} &
  {\color[HTML]{3531FF} \textbf{0.407}} &
  {\color[HTML]{FF0000} \textbf{0.086}} &
  {\color[HTML]{FF0000} \textbf{4.594}} &
  {\color[HTML]{3531FF} \textbf{58.622}} &
  {\color[HTML]{FF0000} \textbf{20.250}} &
  {\color[HTML]{3531FF} \textbf{3.516}} \\ \bottomrule
\end{tabular}}
\end{table*}

\begin{table}[]
\caption{3D detection comparison on the nuScenes nighttime validation set. Both BEVDepth~\cite{li2022bevdepth} and BEVStereo~\cite{li2022bevstereo} are trained using the nuScenes daytime training set. The best and second performance are marked in {\color[HTML]{FE0000} \textbf{red}} and {\color[HTML]{3531FF} \textbf{blue}}. * indicates that it has been retrained on the nuScenes training set.}
\label{tab:detection_result}
\centering
\resizebox{0.48\textwidth}{!}{%

\begin{tabular}{@{}c|cccc|cccc@{}}
\toprule
                          & \multicolumn{4}{c|}{BEVDepth~\cite{li2022bevdepth}}                                 & \multicolumn{4}{c}{BEVStereo~\cite{li2022bevstereo}}                        \\ \cmidrule(l){2-9} 
\multirow{-2}{*}{Methods} & AP$\uparrow$     & ATE$\downarrow$   & ASE$\downarrow$                                  & AOE$\downarrow$   & AP$\uparrow$     & ATE$\downarrow$   & ASE$\downarrow$                          & AOE$\downarrow$   \\ \midrule
Night Image &
  0.134 &
  0.787 &
  {\color[HTML]{000000} 0.195} &
  {\color[HTML]{3531FF} \textbf{0.957}} &
  0.124 &
  0.746 &
  {\color[HTML]{3531FF} \textbf{0.205}} &
  {\color[HTML]{000000} 0.714} \\ \midrule
SCI-diffcult~\cite{ma2022toward}              & 0.067 & 0.828 & 0.187                                 & 1.071 & 0.032 & 0.764 & 0.239                        & 0.774 \\
Zero-DCE++~\cite{li2021learning}                & 0.089 & 0.826 & 0.197                                 & 1.029 & 0.077 & 0.780  & 0.224                        & 0.787 \\
URetinex-Net~\cite{wu2022uretinex}              & 0.053 & 0.831 & 0.184                                 & 1.114 & 0.035 & 0.782 & 0.243                        & 0.803 \\
ExposureDiffusion~\cite{wang2023exposurediffusion}              & 0.040 & 0.829 & {\color[HTML]{FF0000} \textbf{0.179}}                                 & 1.092 & 0.035 & 0.796 & 0.244                        & 0.769 \\
ShadowDiffusion~\cite{guo2023shadowdiffusion}              & 0.072 & 0.851 & 0.184                                 & 1.242 & 0.082 & 0.789 & 0.224                        & 0.718 \\
SNR-Aware*~\cite{xu2022snr}              & 0.089 & 0.817 & 0.193                                 & 1.088 & 0.072 & 0.773 & 0.251                        & 0.742 \\
SCI*~\cite{ma2022toward}                       & 0.062 & 0.829 & 0.181 & 1.133 & 0.034 & 0.783 & {\color[HTML]{000000} 0.235} & 0.801 \\
EnlightenGAN*~\cite{jiang2021enlightengan} &
  {\color[HTML]{3531FF} \textbf{0.138}} &
  {\color[HTML]{3531FF} \textbf{0.786}} &
  {\color[HTML]{000000} 0.193} &
  {\color[HTML]{FF0000} \textbf{0.948}} &
  {\color[HTML]{3531FF} \textbf{0.128}} &
  {\color[HTML]{3531FF} \textbf{0.743}} &
  {\color[HTML]{FF0000} \textbf{0.204}} &
  {\color[HTML]{3531FF} \textbf{0.680}} \\
CLIP\_LIT*~\cite{liang2023iterative}                 & 0.131 & 0.791 & 0.199                                 & 0.972 & 0.121 & 0.753 & 0.211                        & 0.739 \\
Ours &
  {\color[HTML]{FF0000} \textbf{0.176}} &
  {\color[HTML]{FF0000} \textbf{0.774}} &
  {\color[HTML]{3531FF} \textbf{0.180}} &
  {\color[HTML]{000000} 1.108} &
  {\color[HTML]{FF0000} \textbf{0.170}} &
  {\color[HTML]{FF0000} \textbf{0.690}} &
  0.210 &
  {\color[HTML]{FF0000} \textbf{0.620}} \\ \bottomrule
\end{tabular}%
}
\end{table}

\section{Experiments}~\label{Sec:Experiments}

\subsection{Experimental Setup}

\noindent \textbf{Datasets.} To explore the low light enhancement for visual perception tasks on autonomous driving, we conduct experiments on the nuScenes dataset~\cite{caesar2020nuscenes}, which is one of the most popular autonomous driving datasets for multiple visual tasks. It consists of 700 scenes for training, 150 scenes for validation, and 150 scenes for testing. For each scene, it provides images with a resolution of $1,600 \times 900$ from 6 surrounding cameras (front, front left, front right, back, back left, back right) to cover the whole viewpoint, and a 360$^{\circ}$ LiDAR point cloud. Camera matrices including both intrinsic and extrinsic are provided, which establish a one-to-one correspondence between each 3D point and the 2D image plane.
We select all 616 daytime scenes of the nuScenes training set containing total 24,745 camera front images as our training set.
All 15 nighttime scenes in the nuScenes validation set containing total 602 camera front images are as our testing set.

\noindent \textbf{Evaluation Metrics.} We evaluate the low-light enhancement and 3D detection tasks in our experiments. For the quantitative assessment on low-light enhancement task, due to lack of paired day-night data in the real autonomous driving scenario, we employ nine no-reference image quality evaluation (IQA) metrics, MUSIQ~\cite{ke2021musiq}, NIQE~\cite{mittal2012making}, HyperIQA~\cite{su2020blindly},  ILNIQE~\cite{zhang2015feature}, MANIQA~\cite{yang2022maniqa}, NIMA~\cite{talebi2018nima} and TReS~\cite{golestaneh2022no}. 
In 3D perception task, we select the ``Car" category as main object to report Average Precision (AP), along with Average Translation Error (ATE), Average Scale Error (ASE), Average Orientation Error (AOE) in experiments.

\noindent \textbf{Training.}
We deploy the Training Data Generation method as described in Sec.~\ref{sec:data_generation} on the nuScenes daytime training set to obtain the triple modality paired data: 1) an instruction prompt, 2) a trustworthy depth map with LiDAR point cloud  projection, and 3) a degraded dark image. 
We implement our LightDiff for 100 epochs with a batch size of 4 on a
single NVIDIA RTX A6000 GPU. We utilize Adam as the optimizer with the learning rate of $1 \times {10}^{-5}$. Following the setting~\cite{zhang2023adding,rombach2022high}, we resize the input images and condition maps to $512 \time 512$ and adapt the pre-trained SD model~\cite{rombach2022high} with the version of 2.1. To obtain the accurate depth map in the inference stage, we train a pre-trained Depth Estimation Network~\cite{ranftl2020towards} based on the daytime and nighttime images and corresponding LiDAR point cloud projection of nuScenes training dataset.

\noindent \textbf{Inference.}
Given the night images of the nuScenes validation set,
different from the training stage utilizing the real LiDAR point clouds projection to help construct the estimated depth maps, we generate depth maps by a pre-trained depth estimation network. In addition, we apply our proposed Recurrent Lighting Inference (ReLI) to optimize their corresponding text prompts and depth maps.

\noindent \textbf{Comparison Methods.} In our experiments, we compare the generation quality, and 3D detection performance of our proposed LightDiff with other existing representative dark image enhancement related methods.
We evaluate our approach by comparing it with prominent methods including supervised enhancement methods like Afifi et al.~\cite{afifi2021learning}, URetinex-Net~\cite{wu2022uretinex}, SNR-Aware-LOLv1~\cite{xu2022snr}, unsupervised enhancement methods like EnlightenGAN~\cite{jiang2021enlightengan}, CLIP-LIT~\cite{liang2023iterative}, Zero-DCE++~\cite{li2021learning}, and diffusion-based methods like ShadowDiffusion~\cite{guo2023shadowdiffusion}, ExposureDiffusion~\cite{wang2023exposurediffusion}. Some methods have released their pretrained models on different datasets. In order to substantiate the exceptional performance of our approach impartially, we engage in a comparative analysis with these pretrained models. Moreover, we furnish the performance evaluations of retrained unsupervised methods, executed on the identical training set as our methodology, thereby contributing to a more comprehensive validation.

\begin{figure}[htb]
        \centering
        \subfloat{%
        \includegraphics[width=0.98\columnwidth]{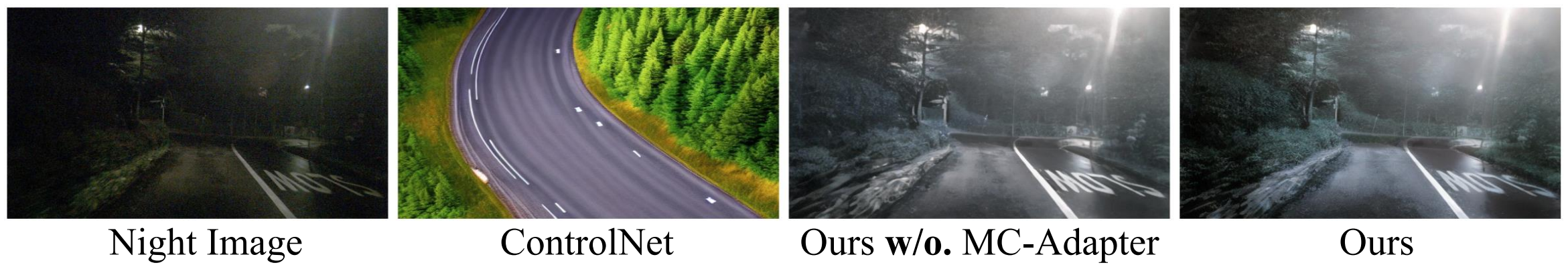}%
        }
    \caption{Visual showcase of our LightDiff with and without the Multi-Condition Adapter. The input for ControlNet~\cite{zhang2023adding} remains consistent, comprising the same text prompt and depth map. Multi-Condition Adapter makes better color contrast and richer details during enhancement.}
	\label{fig:vis_MCA}  
\end{figure}

\subsection{Comparison Results}
\noindent \textbf{Visual comparison.}
We present visual comparisons of some samples from the nuScenes nighttime validation set in Fig.~\ref{fig:comparison_generating}. 
Our method consistently produces visually pleasing results with improved color and eliminated noise.
Moreover, our method excels in handling challenging dark regions, restoring clear texture details and satisfactory luminance without introducing any noise, while other methods may either fail to address such dark regions or produce unsatisfactory results with visible noise.
Specifically,
we can see that compared to RUAS-LOL~\cite{liu2021retinex} and SCI-difficult~\cite{ma2022toward}, our method produces results without over-exposure or under-exposure. 
Our results exhibit better color contrast and input-output consistency in global regions.

\noindent \textbf{Quantitative Comparison.}
It is unattainable to collect nighttime-daytime paired images in real dynamic driving scenarios, currently we rely on several non-reference image quality evaluation (IQA) metric to evaluate the quantitative results. The quantitative comparison on the nuScenes nighttime validation set is presented in Table~\ref{tab:vis_result}.
Our method achieve the best performance in the four no-reference IQA metrics when compared to other methods, demonstrating the satisfactory image quality of our results.
 \\
\noindent \textbf{3D perception Comparison and Visualization.}
For 3D perception task, we only enhance the front camera view of the nuScene nighttime validation set, while other five camera-view are keep the original darkness. 
We utilize two 3D perception state-of-the-art methods BEVDepth~\cite{li2022bevdepth} and BEVStereo~\cite{li2022bevstereo} trained on the nuScenes dayitme training set, which is more effective to collect and annotate in the real-world driving scenario to evaluate the car detection with our effect of generated qualify on perception performance.
We show the quantitative comparison of 3D perception performance on the nuScenes nighttime validation set in Table~\ref{tab:detection_result}. Compared to results on the original nighttime images, by applied our enhanced images, the BEVDepth and BEVStereo can achieve $17.6\%$ AP and $17.0\%$ AP, respectivley, which have the improvement of $4.2\%$ AP and $4.6\%$ AP.
Without any extra training requirement, our proposed method can improve the perception performance for current models by directly applied our generated enhanced images. But some comparison enhancement methods like SCI~\cite{ma2022toward} and Zero-DCE++~\cite{li2021learning}, show the negative effect on 3D perception performance to make a performance drop. We visualize some 3D detection results on front camera-view and  Bird’s-Eye-View (BEV) in Fig.~\ref{fig:comparison_detection}. Our proposed LightDiff not only help the driver see more clear in the darkness, but also  help the deep learning perception detect  more accurately in the challenging real-world dark conditions.

\begin{figure}[htb]
  \centering
  \includegraphics[width=0.95\linewidth]{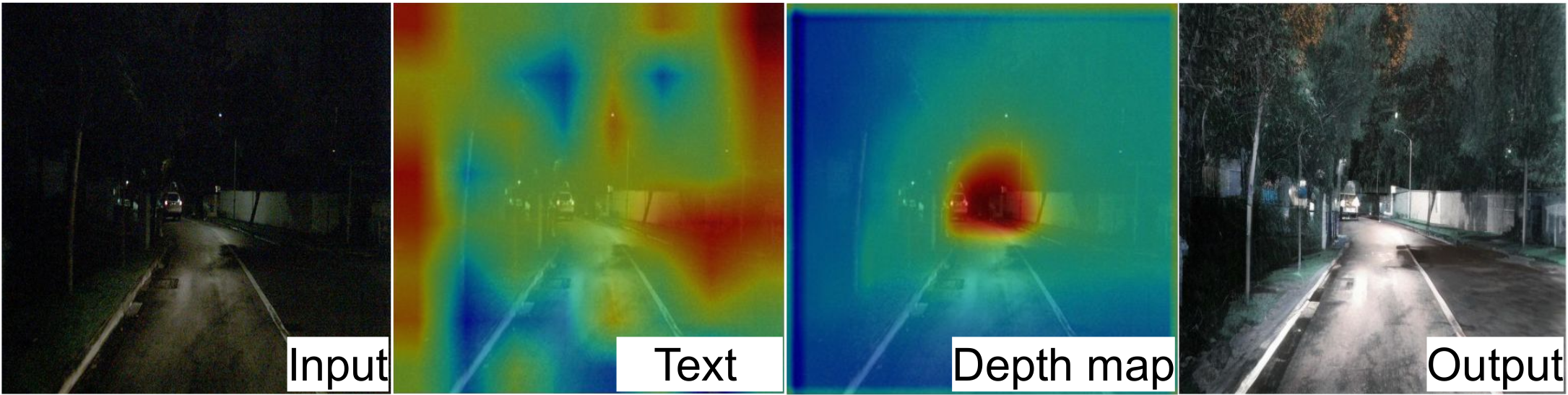}
   \caption{The examples of attention map for different modality inputs.}
   \label{attmap}
\end{figure}

\begin{figure}[htb]
	\begin{minipage}[b]{1\columnwidth}
		\centering
		\includegraphics[width=1\columnwidth]{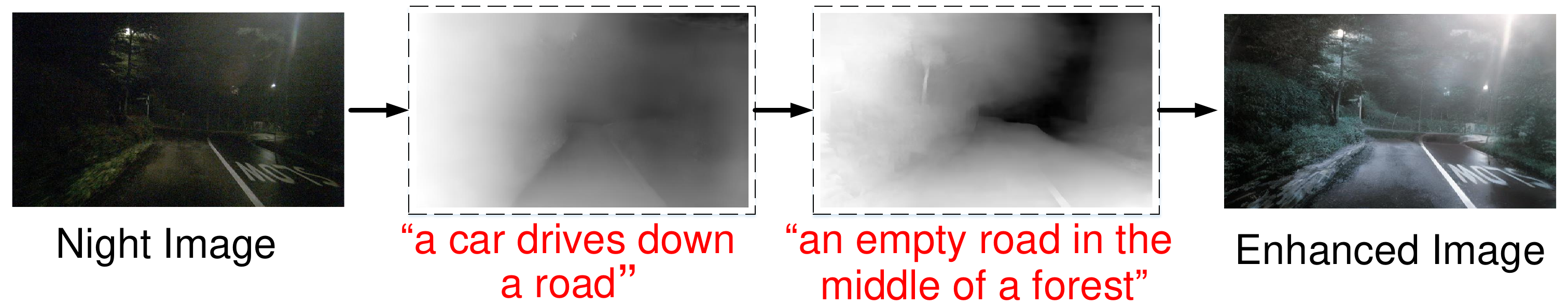}
	\end{minipage}
    \caption{Illustration of enhanced multi-modality generation through Recurrent Lighting Inference (ReLI). Improved  accuracy in text prompts and depth map prediction are achieved by invoking the ReLI once.}
	\label{fig:enhanced modality} 
\end{figure}

\vspace{-0.5em}
\subsection{Ablation Study}

\begin{table}[htb]
\caption{Ablation study for multi-modality inputs on the nuScenes nighttime validation set. $*$ indicates that it has been retrained on the nuScenes training set.}
\centering
\resizebox{1.0\columnwidth}{!}{%
\begin{tabular}{@{}c|cc|cccc@{}}
\toprule
 &
  \multicolumn{2}{c|}{3D Detection} &
  \multicolumn{4}{c}{Dark Enhancement} \\ \cmidrule(l){2-7} 
\multirow{-2}{*}{Modality Input} &
  \multicolumn{1}{c|}{AP$\uparrow$} &
  ASE$\downarrow$ &
  \multicolumn{1}{c|}{MUSIQ$\uparrow$} &
  \multicolumn{1}{c|}{TReS$\uparrow$} &
  \multicolumn{1}{c|}{ILNIQE$\downarrow$} &
  NIQE$\downarrow$ \\ \midrule
Image-only &
  \multicolumn{1}{c|}{0.119} &
  0.190 &
  \multicolumn{1}{c|}{40.103} &
  \multicolumn{1}{c|}{30.472} &
  \multicolumn{1}{c|}{36.789} &
  5.717 \\
\textbf{w/o.} Text  &
  \multicolumn{1}{c|}{0.146} &
  0.191 &
  \multicolumn{1}{c|}{42.380} &
  \multicolumn{1}{c|}{31.367} &
  \multicolumn{1}{c|}{35.761} &
  5.650 \\
\textbf{w/o.} Depth Map &
  \multicolumn{1}{c|}{0.139} &
  { 0.188} &
  \multicolumn{1}{c|}{43.400} &
  \multicolumn{1}{c|}{33.504} &
  \multicolumn{1}{c|}{32.974} &
  5.419 \\
Image-only$^{*}$ &
  \multicolumn{1}{c|}{0.132} &
  0.182 &
  \multicolumn{1}{c|}{44.766} &
  \multicolumn{1}{c|}{49.304} &
  \multicolumn{1}{c|}{23.471} &
  4.353 \\
\textbf{w/o.} Text$^{*}$  &
  \multicolumn{1}{c|}{0.165} &
  0.183
   &
  \multicolumn{1}{c|}{47.700} &
  \multicolumn{1}{c|}{36.723} &
  \multicolumn{1}{c|}{30.064} &
  5.144 \\
\textbf{w/o.} Depth Map$^{*}$ &
  \multicolumn{1}{c|}{0.122} &
  \textbf{0.174} &
  \multicolumn{1}{c|}{47.286} &
  \multicolumn{1}{c|}{43.340} &
  \multicolumn{1}{c|}{21.655} &
  \textbf{3.358} \\
Ours &
  \multicolumn{1}{c|}{ \textbf{0.176}} &
   0.180 &
  \multicolumn{1}{c|}{\textbf{51.674}} &
  \multicolumn{1}{c|}{\textbf{58.622}} &
  \multicolumn{1}{c|}{ \textbf{20.250}}&
   3.516 \\ \bottomrule
\end{tabular}%
}
\label{tab:aba_modality}
\end{table}

\begin{table}[]
\caption{Ablation study for each proposed component on the nuScenes nighttime validation set.}
\label{tab:ablation}
\resizebox{\columnwidth}{!}{%
\begin{tabular}{@{}c|cc|cccc@{}}
\toprule
 &
  \multicolumn{2}{c|}{3D Detection} &
  \multicolumn{4}{c}{Dark Enhancement} \\ \cmidrule(l){2-7} 
\multirow{-2}{*}{} &
  \multicolumn{1}{c|}{AP$\uparrow$} &
  ASE$\downarrow$ &
  \multicolumn{1}{c|}{MUSIQ$\uparrow$} &
  \multicolumn{1}{c|}{TReS$\uparrow$} &
  \multicolumn{1}{c|}{ILNIQE$\downarrow$} &
  NIQE$\downarrow$ \\ \midrule
\textbf{w/o.} LDRM &
  \multicolumn{1}{c|}{0.151} &
  0.197 &
  \multicolumn{1}{c|}{41.387} &
  \multicolumn{1}{c|}{34.056} &
  \multicolumn{1}{c|}{{\color[HTML]{000000} 25.353}} &
  3.569 \\
\textbf{w/o.} MC-Adapter &
  \multicolumn{1}{c|}{0.152} &
  0.183 &
  \multicolumn{1}{c|}{47.370} &
  \multicolumn{1}{c|}{52.342} &
  \multicolumn{1}{c|}{21.666} &
  3.841 \\
\textbf{w/o.} ReLI &
  \multicolumn{1}{c|}{0.163} &
  {\color[HTML]{000000} 0.184} &
  \multicolumn{1}{c|}{48.277} &
  \multicolumn{1}{c|}{54.462} &
  \multicolumn{1}{c|}{22.423} &
  4.085 \\
Ours &
  \multicolumn{1}{c|}{ \textbf{0.176}} &
   \textbf{0.180} &
  \multicolumn{1}{c|}{\textbf{51.674}} &
  \multicolumn{1}{c|}{\textbf{58.622}} &
  \multicolumn{1}{c|}{ \textbf{20.250}}&
   \textbf{3.516} \\ \bottomrule
\end{tabular}%
}
\end{table}

To validate the effectiveness of our proposed components, we provide the quantitative comparison on 3D perception and dark enhancement tasks in Table~\ref{tab:ablation}. The visual comparison results of  Fig.~\ref{fig:vis_MCA} show the effectiveness of discerning the significance of different visual conditions.
The heatmaps in Fig.~\ref{attmap} illustrate the correlation of each image pixel with the two different modality inputs.
The Table~\ref{tab:aba_modality} unequivocally demonstrates the beneficial impact of each modality input within our LightDiff.
We present the effectiveness of Recurrent Lighting Inference (ReLI), which can optimize the accuracy of multi-modality generation effectively in Fig.~\ref{fig:enhanced modality}. It indicates that our LightDiff can produce better color contrast and richer details with our Multi-Conditional Adapter. 
The result in Table~\ref{tab:ablation} clearly justifies the positive effect of each proposed component of our LightDiff.

\vspace{-0.5em}
\section{Conclusions}\label{Sec:Conclusions}

This paper introduces LightDiff, a domain-tailored framework designed to enhance the low-light image quality for autonomous driving applications, mitigating the challenges faced by vision-centric perception systems. By leveraging a dynamic data degradation process, a multi-condition adapter for diverse input modalities, and perception-specific score guided reward modeling using reinforcement learning, LightDiff significantly enhances the image quality and 3D vehicle detection in nighttime on the nuScenes dataset. This innovation not only eliminates the need for extensive nighttime data but also ensures semantic integrity in image transformation, demonstrating its potential to enhance safety and reliability in autonomous driving scenarios. Without realistic paired day-night images, synthesizing dark driving images with vehicle lights is quite difficult, limiting the research in this field. Future research can be focused on a better collection or generation of the high-quality training data. 
\\
\noindent \textbf{Acknowledgments:} This work is supported in part by NSF 2215388, and the National Research Foundation, Singapore, and DSO National Laboratories under the AI Singapore Programme (No: AISG2-GC-2023-008).

{
    \small
    \bibliographystyle{ieeenat_fullname}
    \bibliography{jinlong}
}


\end{document}